\documentclass{article}
\usepackage{ijcai21}

\usepackage{times}

\usepackage{soul}
\usepackage{url}
\usepackage[hidelinks]{hyperref}
\usepackage[utf8]{inputenc}
\usepackage[small]{caption}
\usepackage{graphicx}
\usepackage{amsmath}
\usepackage{booktabs}
\usepackage{courier}
\usepackage{xcolor}
\usepackage{xspace}

\graphicspath{{./images/}}

\newcommand{\sys}{HAFLO\xspace}

\title{\sys: GPU-Based Acceleration for Federated Logistic Regression}

\author{
Xiaodian Cheng$^1$\and
Wanhang Lu$^2$\and
Xinyang Huang$^2$\and
Shuihai Hu$^2$\And
Kai Chen$^{1,3}$\\
\affiliations
$^1$iSING Lab, The Hong Kong University of Science and Technology\\
$^2$Clustar\\
$^3$Peng Cheng Lab\\
\emails
xchengaq@connect.ust.hk,
wanhang.lu@clustar.ai,
minel.huang@clustar.ai,
shuihai@clustar.ai,
kaichen@cse.ust.hk
}

\begin{document}

\maketitle

\begin{abstract}
In recent years, federated learning (FL) has been widely applied for supporting decentralized collaborative learning scenarios. Among existing FL models, federated logistic regression (FLR) is a widely used statistic model and has been used in various industries. To ensure data security and user privacy, FLR leverages homomorphic encryption (HE) to protect the exchanged data among different collaborative parties. However, HE introduces significant computational overhead (i.e., the cost of data encryption/decryption and calculation over encrypted data), which eventually becomes the performance bottleneck of the whole system. In this paper, we propose \sys, a GPU-based solution to improve the performance of FLR. The core idea of \sys is to summarize a set of performance-critical homomorphic operators (HO) used by FLR and accelerate the execution of these operators through a joint optimization of storage, IO, and computation. The preliminary results show that our acceleration on FATE, a popular FL framework, achieves a 49.9$\times$ speedup for heterogeneous LR and 88.4$\times$ for homogeneous LR.
\end{abstract}

\section{Introduction}
The development of distributed computing and machine learning provides various industries like internet, finance and medicine with great support. However, with increasing concerns on data leakage and the introduction of relevant laws and regulations, enterprises and institutions are strictly restricted from transacting internal data with others. Thus, the data held by different parties is isolated and the value of data cannot be fully utilized. This problem is known as \emph{data island problem}~\cite{yang2019federated}. In recent years, there is an increasing interest in applying privacy computing to solve the data island problem. Several privacy computing solutions have been proposed, including Secure Multi-party Computation (MPC)~\cite{goldreich1998secure}, Trusted Execution Environment (TEE)~\cite{ohrimenko2016oblivious}, Federated Learning (FL)~\cite{konevcny2016federated}, etc. In the area of multi-party collaborative learning, FL is the most promising technique. The most distinctive idea of FL is that during the process of distributed model building, the datasets are only accessible by local devices without being shared among different parties, which strongly guarantees the protection for data privacy. Currently, federated logistic regression (FLR) is one of the most widely applied federated models, and has been implemented in almost all the FL frameworks.

A number of alternative security techniques can be used to ensure data privacy~\cite{yang2019federatedbook}, which includes Homomorphic Encryption (HE)~\cite{rivest1978data}, Oblivious Transfer (OT)~\cite{rabin2005exchange}, Garbled Circuit (GC)~\cite{yao1986generate}, Differential Privacy (DP)~\cite{dwork2006calibrating}, etc. Due to its ability to allow calculations over ciphertext directly, HE has been widely applied in existing industrial implementations of FLR. With HE, different participants in FL are able to achieve collaborative modeling through exchanging encrypted parameters instead of user data or models. However, the computational complexity of HE introduces significant overhead in FLR, and the cost of homomorphic computation eventually becomes the performance bottleneck in the model training process. Most existing FL frameworks, such as FATE~\cite{webank19}, purely rely on CPU to accomplish huge amounts of homomorphic computation. However, the computing power of CPU is quite limited, even when multiple cores are utilized concurrently.

Unlike CPU, various specifically designed hardware processors are suitable for handling operations with large amounts of parallelism. In recent years, the use of heterogeneous system which combines CPU with hardware accelerators becomes a popular choice in scenarios where high computing throughput is required. An encryption framework based on Field Programmable Gate Array (FPGA) was proposed in~\cite{yang2020fpga} for computational acceleration. In addition, Graphics Processing Unit (GPU) is widely used to offload the computational workload from CPU because of its high instruction throughput and programming flexibility.

Following this trend, we aim to design a GPU-CPU heterogeneous system to accelerate the homomorphic computation of FLR. The calculation of encrypted gradient descent and loss function in FLR are algorithms with high computational complexity. They are executed in every iteration of the training process until convergence is achieved. These complicated and repetitive calculations are good matches for GPU acceleration.

%Therefore, our application is able to provide huge improvement on performance with high operability.

Despite the great opportunities of GPU-based heterogeneous system, there remains several challenges to overcome. First, the algorithms of FLR are complex, involving very diverse arithmetic expressions. Hence it is difficult to directly offload all the computation tasks onto GPU. Second, the discrete data storage and sequential workflow of existing FL frameworks make it challenging for GPU to perform large-batch parallel computation. Third, frequent data copy and I/O transfer between different devices will introduce additional delays.

%The difference of the data structure between FL frameworks and GPU makes it difficult to achieve high compatibility. In addition, Frequent data copies and transfers introduced by external device will greatly affect the performance of the system. We will detail the obstacles in section~\ref{sec:challenges}.

Motivated by above challenges, we propose \sys, a GPU-based acceleration solution for FLR. To achieve high performance and reduce the complexity of computation offloading, we summarize a set of performance-critical homomorphic operators (HO) used by FLR, and do a joint optimization of storage, IO and computation to accelerate the execution of these operators on GPU.

%For the purpose of low high compatibility, GPU is assigned to handle operations with finer granularity. Aiming at high performance, we try to respectively minimize the number of data format conversion and data transaction between GPU and CPU.

Our contributions are summarized as follows:
\begin{itemize}
    \item \emph{Storage optimization.} We propose an aggregated data storage system to achieve high computation efficiency on GPU, including data format conversion and storage management.
    \item \emph{Computation optimization.} Instead of directly accelerating original complex algorithms, we summarize a set of performance-critical HOs used by FLR which are of lower implementation complexity and much easier for GPU acceleration.
    \item \emph{IO optimization.} We propose a scheme to temporarily store intermediate results in GPU memory, and implement corresponding storage management processes which can be controlled by upper frameworks to reduce the transactions of data between GPU and CPU.
\end{itemize}

The rest of this paper is organized as follows. In section~\ref{sec:background}, we introduce necessary background about FLR and applications of GPU in general-purpose computing. We also analysis the challenges in applying GPU acceleration in FLR. In section \ref{sec:design}, we describe the design of our acceleration scheme in detail. Finally, we present the preliminary testing results of our implementation in section~\ref{sec:result} and draw a conclusion in section~\ref{sec:conclusion}.

\section{Background}\label{sec:background}
\subsection{Federated Logistic Regression}\label{sec:FLR}
FLR~\cite{hardy2017private} is a deformation of traditional logistic regression. Based on whether datasets in different parties share similar feature space or similar instance ID space, FLR is classified into homogeneous LR and heterogeneous LR. The basic mathematical principles in both models are identical. To protect privacy and keep data operability in FLR, Homomorphic Encryption (HE) is adapted for data encryption before remote parameter sharing. Paillier Cryptosystem~\cite{paillier1999public}, a pragmatic semi-homomorphic encryption algorithm, is a popular choice for FLR.

In every iteration of training, stochastic gradient descent is performed. The stochastic gradients of logistic regression are
$$ \nabla l_{S^{\prime}}\left({\bf {\theta}}\right)=\frac{1}{s^\prime}\sum_{i\in S^{\prime}}\left(\frac{1}{1+e^{-y{\bf{\theta}}^\top\bf{x}}-1}\right)y_i{\bf{x}}_i, $$ \label{algo1}
where $ S^{\prime} $ is a mini-batch.
The training loss on the dataset is
$$ l_S(\theta)=\frac{1}{n}\sum_{i\in S}\log(1+e^{-y_i{\bf{\theta}}^{\top}{\bf{x}}_i}). $$ \label{algo2}
However, Paillier Cryptosystem limits the type of operations on cipher text to addition. Formulas above do not satisfy the limitation. Therefore, they need to be replaced with their second order approximation by taking Taylor series expansions of exponential functions. After approximation, the gradients and loss function are respectively
$$ \nabla l_{S^{\prime}}\left({\bf {\theta}}\right)\approx\frac{1}{s^{\prime}}\sum_{i\in S^{\prime}}\left(\frac{1}{4}{\bf{\theta}}^{\top}{\bf{x}}_i-\frac{1}{2}y_i\right){\bf{x}}_i, $$ \label{algo3}
and
$$ l_H(\theta)\approx\frac{1}{h}\sum_{i\in H}\log2-\frac{1}{2}y_i{\bf{\theta}}^\top{\bf{x}}_i+\frac{1}{8}\left({\bf{\theta}}^\top{\bf{x}}_i\right)^2. $$ \label{algo4}
Combining the above principles with HE, multiple parties can build FLR models cooperatively.

In a typical FLR workflow, the participants start every iteration from training on their local model separately. During this process, they need to exchange encrypted intermediate data. Afterwards, they transfer encrypted local training results, like local gradients and loss, to a coordinator. The coordinator aggregates ciphertexts from different participants and performs decryption. After decryption. it sends aggregated result back to the participants for them to update local models.

\subsection{GPU Acceleration for AI Framework}
A GPU is a computing device whose architecture is well-suited for parallel workload. It was traditionally used for graphics processing. Emerging programmability and increasing performance make GPU attractive in general-purpose computations, including database queries, scientific computing, etc. Benefiting from hundreds of small processor cores and unique hierarchical storage structure, GPU can execute more arithmetic tasks concurrently than CPU. GPUs have been applied for acceleration in Artificial Intelligence for decades. Open-source machine learning frameworks like TensorFlow~\cite{abadi2016tensorflow} and PyTorch~\cite{paszke2019pytorch} have specialized acceleration for computing with GPU. As the developers become increasingly reliant on open-source frameworks, GPU has become an indispensable part of machine learning.

\subsection{Opportunities and Challenges}\label{sec:challenges}
As a necessary privacy technique, HE inevitably introduces heavy time overhead to FL system. The common key length used today is 1024 bits. It means that original floating-point operations are converted into those among large integers with thousands of bits in length. Moreover, Paillier Cryptosystem~\cite{paillier1999public} has following properties.
$$ E(a+b)=E(a)*E(b)\mod\ n$$
$$ E(a*b)=E(a)^b\mod\ n,$$
where $E$ is the Paillier encryption function.

In common situations, large amounts of calculation instances are fed into the system, which are typical scenarios for data stream processing. 

Under above circumstances, the computing power of CPU is far from sufficient to meet the demands in applications. In contrast, GPU is efficient to handle such tasks with pipeline processing. Calculations among large integers can be divided into multiple stages and thus assigned to different threads for acceleration on GPU. Moreover, multiple instances can be operated concurrently with proper GPU resource allocation in order to maximize the performance.

Therefore, offloading homomorphic calculations onto GPU is a straightforward solution for relieving this bottleneck.

However, the leverage of GPU generates other problems. 

\begin{itemize}
\item To facilitate transactions of data, major FL frameworks apply distributed computation system like Spark~\cite{zaharia2012resilient} to manage basic calculation and communication instructions. To make data in GPU memory compatible with different databases, time-consuming data format conversions and memory access operations will be introduced. Meanwhile, most basic instructions in FL frameworks are only performed on a small amount of data because of instance ID based data partition. In some cases, only data of a single instance maybe handled concurrently. However, GPU follows SIMD (Single Instruction, Multiple Data) architecture. Such operations greatly waste GPU computing resources and data bandwidth, which will end up in catastrophic performance loss.
\item The algorithms in FLR are complex, considering the calculations of various parameters along with encryption and decryption operations. Different FL frameworks use divergent implementations and workflows. It's hard to achieve acceleration with high compatibility without large amounts of work.
\item Along with frequent parameter exchanges in FLR and inefficient bandwidth of physical data link, cross-device data transmission between CPU and GPU leads to non-negligible overheads in the training process.
\end{itemize}

\section{Design}\label{sec:design}
\subsection{Design Overview}
%We propose a heterogeneous system as follows. The architecture of this system is shown in Figure~\ref{eq:Arch_Overview}. It is envisioned to be deployed for every participant involved in FLR based on major FL frameworks. There are two devices in this system, CPU and GPU. In the system, CPU transfers data with GPU through physical interfaces like PCIe or NVLink. In GPU, we implement several GPU kernels to accelerate basic operators, which we detail in section~\ref{sec:operators}. CPU is responsible for the execution of FL frameworks. To efficiently interconnect GPU kernels with FL frameworks, a middleware is designed for data conversion, storage management and operation execution. The middleware is discussed in section~\ref{sec:storage}.

\begin{figure}[!htbp]
    \centering
    \includegraphics[width=0.44\textwidth]{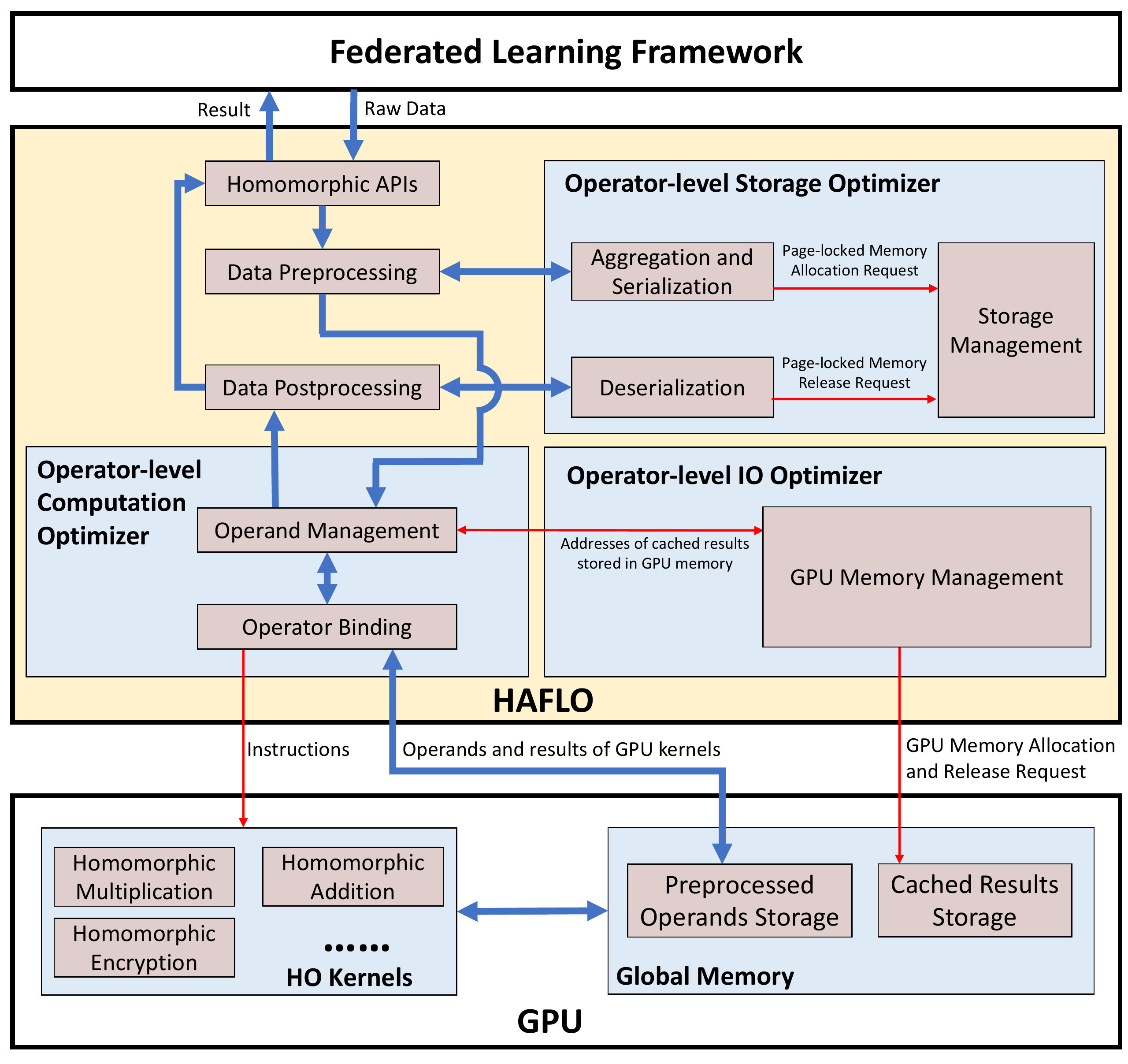}
    \caption{Architecture overview of \sys}
    \label{eq:Arch_Overview}
\end{figure}

We propose \sys, a GPU-based heterogeneous system for accelerating the homomorphic calculations in FLR. As shown in Figure~\ref{eq:Arch_Overview}, \sys mainly has three components, an Operator-level Computation Optimizer for computation optimization, an Operator-level Storage Optimizer for storage optimization and an Operator-level IO Optimizer for IO optimization. In our system, all the operations are at the operator-level of granularity.

\sys provides a set of APIs for supporting upper-layer FL framework to execute operator-level homomorphic calculations over the delivered raw data with GPU. In the first step, the Storage Optimizer will perform data preprocessing including aggregation and serialization over the raw data, which can significantly facilitate the SIMD batch calculation on GPU. After data preprocessing, the operand management module will prepare all the operands needed by following operator calculation. It is possible that the operator calculation will demand the results of previous operator calculations. In this case, the operand management module will also receive the addresses of cached results stored in GPU memory. After all the operands are prepared, the Operator Binding module will interpret task information and generate instructions to invoke corresponding HO kernel(s) for executing the calculation. In the meanwhile, the preprocessed operands will be copied to GPU global memory.

After calculation, Operator Binding reads the results stored in GPU memory and sends them back to Operand Management. Next, Storage Optimizer performs data postprocessing including deserialization over the results before they are returned to FL framework. If required, the developers can also choose to cache the results in the GPU memory by setting the value of the corresponding API parameters. In this case, data will not be copied back to CPU.

\subsection{Operator-level Storage Optimizer}\label{sec:storage}
% In this subsection, we present the design for middleware, which is also a data management system. It is the key system that interconnects GPU with FL frameworks. From a high level, the system generates batched data for GPU and calls GPU kernels in a generic way. To improve efficiency, middleware consists of two major components: Preprocessor and GPU datastore.
Storage Optimizer is responsible for data format conversion from FL framework to GPU operators. In order to optimize data transfer between CPU and GPU as well as utilization of GPU resources, data aggregation and format conversion are necessary for every operator. To improve efficiency, proper management of memory is also indispensable.

\subsubsection{Storage Management}\label{sec:storagemanagement}
Storage Management is the basic component for Storage Optimizer. It can be leveraged to allocate or free data buffer in CPU memory. Specifically, Storage Management assigns a continuous page-locked memory for every allocation request on CPU. Different from pageable memory, page-locked memory can be directly accessed by GPU device, which greatly reduces transferring time.

However, allocation of page-locked memory is quite drawn-out, even slower than cross-device data transfer in some cases. To solve the problem, there are several facts in FLR that are worthwhile to note. Stochastic Gradient Descent (SGD) is a common algorithm for logistic regression. We observe from major frameworks that SGD is performed with the same mini-batch separation in different epochs during training. Moreover, in a training task, most mini-batches have the same batch size. Motivated by above observations, we designed a dynamic memory allocation scheme to reuse the page-locked memory. As long as a new memory space is allocated, its ID, size, availability and offset address will be recorded using a table. When there comes another memory allocation request, Storage Management checks the table for available memory space with the same size. If such space exists, Storage Management changes the status of space to be unavailable and return the address. Similarly, when data stored in a memory space is no longer needed, Storage Management changes the status of this memory space to be available. To optimize the searching progress, we use memory size as the table index.

A garbage collection mechanism is also applied to avoid overloading the system memory with too much redundant space. We use LRU (Least recently used) algorithm to manage memory space. When total size of allocated memory space exceeds range of tolerance, the least recent used memory space is released.

\subsubsection{Aggregation and Serialization}\label{sec:preprocessor}
As we stated in section~\ref{sec:challenges}, most of the calculations in major frameworks are executed over a small amount of data because of data partition. To solve this problem, dataset aggregation is introduced in data preprocessing. The main goal for this component is to reconstruct datasets for efficient batch processing on GPU. In FLR, Storage Optimizer loads raw data from database and densely packs the features and labels which come from the same mini-batch together in an array-like architecture. As discussed before, the separation of mini-batch remains the same in different epochs. On basis of this workflow, we only need to execute aggregation once after the creation of mini-batches since aggregated datasets can be reused in following epochs. 

After aggregation, serialization is performed to convert data format. In order to optimize the performance of GPU and fully utilize bandwidth between CPU and GPU, data structure should be converted into byte stream before it is sent to GPU. To convert data into byte stream, Storage Optimizer performs data rearrangement to get serialized data stored in page-locked memory space preassigned by Storage Management. After serialization, byte stream is forwarded to subsequent modules for calculation.

\subsubsection{Data Postprocessing}\label{sec:postprocessor}
When all the calculations are completed, deserialization is necessary to recover data for FL frameworks, such as the recovery of overall gradient obtained after decryption. This is a reverse process of serialization. Calculation results are recreated from byte stream stored in page-locked memory space. After data recovery, Storage Management is requested to free up the memory.

\subsection{Operator-level Computation Optimizer}\label{sec:operators}
After data preprocessing, operator calculation is ready for execution. \sys mainly has two computation optimizations. First, several high-performance GPU-based operators are built to accelerate the process of arithmetic computation. Second, we introduce an operand management scheme to reuse intermediate data temporarily cached in GPU memory, which effectively reduces memory copy and cross-device data transfer.

\subsubsection{Operator Binding}
As mentioned in section~\ref{sec:challenges}, given the complex computing process and divergent implementations, it's infeasible to individually accelerate every calculation task with GPU. In addition, some computations are not cost-effective to be handled by GPU due to their low computational complexity. According to the principles and workflow of FLR described in section~\ref{sec:FLR}, the most time-consuming parts in training process are homomorphic calculations over encrypted intermediate data. With Taylor Approximation, such calculations in FLR are mainly composed of addition and multiplication. Combined with the properties of HE, we can sort out all the basic operators required.

Therefore, we summarize a set of performance-critical HOs with high computational complexity from the FLR algorithms and build corresponding GPU kernels for their acceleration. Examples of implemented operator kernels are listed in Table~\ref{tab:operators}. For every kernel in our design, multiple computing cores are utilized for parallel calculation. Considering the size of on-board memory in modern GPUs, more than 10 million homomorphic tasks can be handled simultaneously. These highly efficient kernels can easily component most instructions during training, including calculations of gradient and loss.

\begin{table}[!htbp]
\centering
\resizebox{0.48\textwidth}{!}{%
\begin{tabular}{ll}
\hline
\multicolumn{1}{c}{\textbf{Operator Names}} & \multicolumn{1}{c}{\textbf{Mathematical formulas}} \\ \hline
Homomorphic Encryption & $ E(m) =g^m*r^n\mod\ n^2 $. \\ \hline
Homomorphic Multiplication & $ E(m_1*m_2)=[E(m_1)]^{m_2}\mod\ n^2 $ \\ \hline
Homomorphic Addition & $ E(m_1+m_2)=E(m_1)*E(m_2)\mod\ n^2 $ \\ \hline
Homomorphic Summation & $ E(\sum m_i)=\prod E(m_i)\mod\ n^2 $ \\ \hline
\end{tabular}%
}
\caption{Examples of basic operators implemented on GPU. $E$ is encryption function. $n$ and $g$ are the public keys.}
\label{tab:operators}
\end{table}

Based on HO kernels, we designed Operator Binding which transfers data with GPU and invokes HO kernel(s) according to the task configurations it receives. The execution of Operator Binding achieves high throughput rate with thorough GPU resource allocation and workflow management. Specifically, the number of threads and blocks allocated for each kernel matches the feature of large integers in homomorphic calculations. Meanwhile, we increase parallelism by overlapping GPU operations, CPU operations and data transmission. 
% We illustrate the combination of operators by an example of the gradient calculation process.

% \subsubsection{Homomorphic APIs}
% We provide Homomorphic APIs which can be used by FL frameworks to invoke GPU kernels with considerable flexibility. The APIs are mostly implemented by overloading operators. As a result, they can be called implicitly without much modification to FL frameworks. 

\subsubsection{Operand Management}\label{sec:operand_management}
Operand Management is designed to prepare data for operator execution. In the most basic workflow, Operand Management receives serialized data from Preprocessor and passes it to Operator Binding with task information. However, in certain cases, Operand Management can reduce efficiency loss with the cooperation of IO Optimizer. It is mentioned in section~\ref{sec:challenges} that data bandwidth between CPU and GPU is limited and could become the performance bottleneck. In addition, serialization and deserialization of ciphertexts conducted in Storage Optimizer are expensive considering the large amounts of memory operations. Hence it is important to reduce above operations.

We note that in the workflow of FLR, there are several consecutive operations whose calculation results are just required by the next few steps. Therefore, aimed at avoiding unnecessary cross-device data IO and data format conversion for every single operator calculation, \sys allows intermediate results to be temporarily stored in GPU memory until they are no longer needed by subsequent operator calculation(s). In such cases, GPU memory addresses of data should also be treated as the operands in \sys. We will detail the complete workflow in section~\ref{sec:IO}. 

Due to the existence of above operations, Operand Management is required to identify and transfer operands with different data types. In addition, for each piece of data stored in GPU memory, Operand Management constructs a data structure which contains useful information, including memory address, data size, etc. This structure is returned to FL framework after calculation for convenient information preservation in the training process. Conversely, Operand Management will extract the information of data and pass it to Operator Binding when FL framework sends the data structure back for subsequent calculations.

\subsection{IO-Optimized Computing Workflow}\label{sec:IO}
In section~\ref{sec:operand_management}, we mentioned the scheme of caching intermediate results in GPU memory. IO Optimizer is designed to manage GPU memory space for related operations. It provides a flexible and effective scheme for caching results in GPU memory and thus significantly reduces the delay spent on data IO.

% The main challenges for such implementation are memory constraint and complexity of arithmetic computations. 

\subsubsection{GPU Memory Management}
For operators without data dependency on results of previous operator(s), GPU memory is allocated directly by Operator Binding to store the required data. However, for operators with data dependency, in order to fetch the cached results for the calculation of subsequent operators, the addresses are exposed to FL framework. This will lead to potential problems including memory leak or invalid access. To ensure memory safety, we use an isolated module (i.e. GPU Memory Management) to manage GPU memory space for caching results. Based on the requirements of training process, GPU Memory Management is called by Operand Management to initialize GPU memory allocation and release requests for cached calculation results. For clarity, we differentiate with GPU memory space used to store preprocessed operands (i.e., Preprocessed Operands Storage in Figure~\ref{eq:Arch_Overview}) and those utilized to cache intermediate results (i.e., Cached Results Storage in Figure~\ref{eq:Arch_Overview}).

\subsubsection{Workflow of temporary storage in GPU memory}
\begin{figure}[!htbp]
    \centering
    \includegraphics[width=0.42\textwidth]{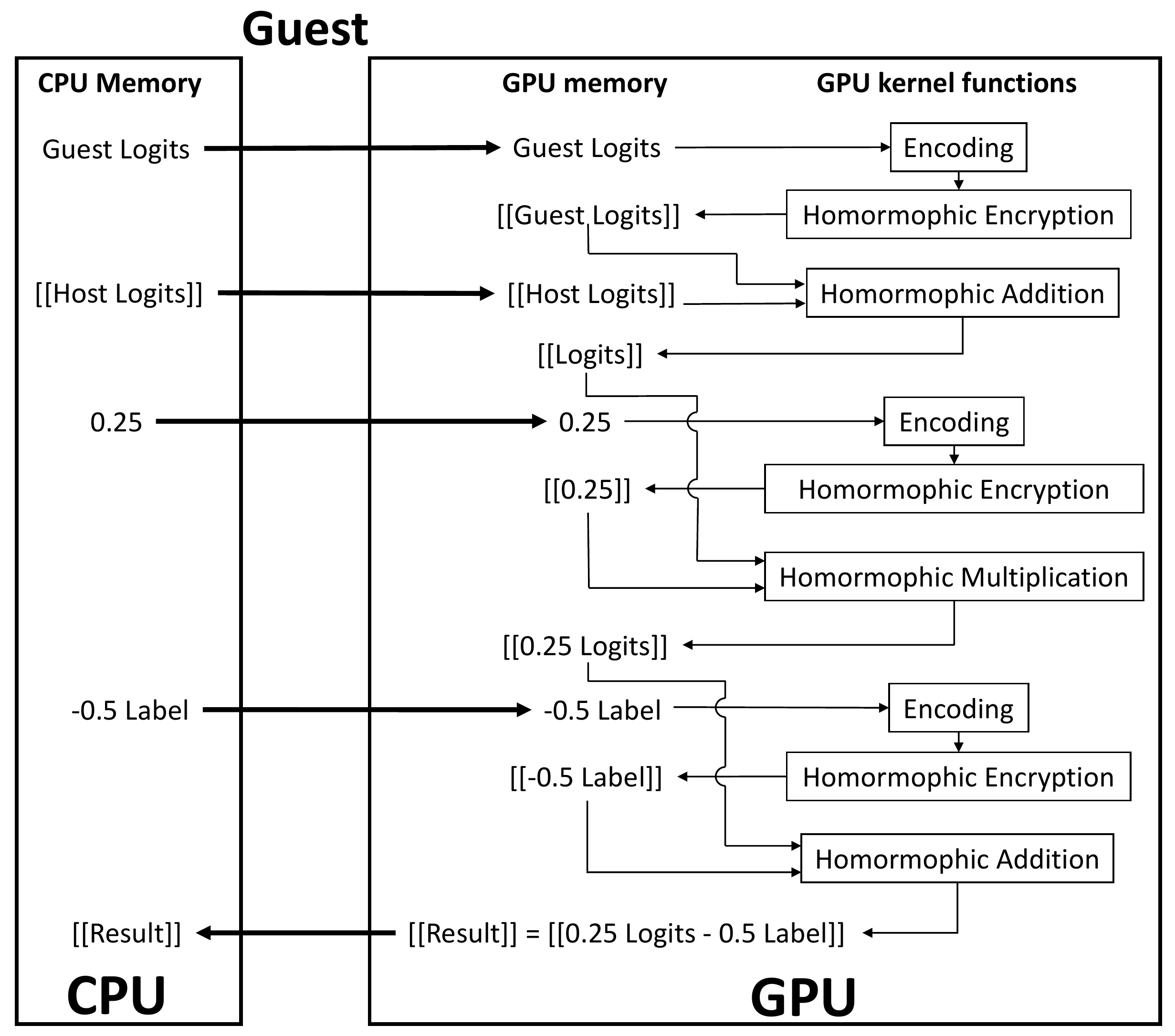}
    \caption{The workflow for guest to calculate $ [[fore\_gradient]] = [[0.25 * (logits_{host} + logits_{guest}) - 0.5 * label]]$, where $[[x]]$ represents corresponding ciphertexts of data $x$. The texts with boxes represent kernel functions. Other texts represent data.}
    \label{eq:Steaming_Flow}.
\end{figure}

In order to reuse intermediate results cached in GPU memory, Operand Management cooperates with GPU Memory Management according to task information. Following the properties of FLR algorithms, we modify the instructions in FL framework with proper arrangement of data storage and transmission, like the example presented in Figure~\ref{eq:Steaming_Flow}. In this example, guest computes fore gradient with the minimum data transaction. In the calculation process, \sys needs to perform 8 basic operators under the instructions of FL framework. Each of the operators follows the same work process below.

\begin{itemize}
    \item If results are required to be cached in GPU memory, Operand Management leverages GPU Memory Management to request memory allocation on GPU. Then it sends the address of allocated memory space to Operator Binding for storing results. After calculation, the address will be packed into a data structure containing data information and returned to FL framework as an intermediate operation result. In such operations, Data Postprocessing is no longer necessary and it can be bypassed.
    \item If some operand(s) in current operation are previous intermediate results which are already cached in GPU memory, FL framework is required to input the data structure containing corresponding information via Homomorphic APIs. Because the data can be directly accessed by GPU, Preprocessing is skipped similarly. Operand Management retrieves data information including memory addresses from data structure and sends it to Operator Binding. If the data is no longer needed in following operations, Operand Management will leverage GPU Memory Management to release memory space after current calculation.
\end{itemize}

\begin{table*}[!htbp]
\centering
\resizebox{\textwidth}{!}{%
\begin{tabular}{lccc}
\hline
\multicolumn{1}{c}{\textbf{Operators}} & \multicolumn{1}{c}{\textbf{\begin{tabular}[c]{@{}c@{}}Throughput of GPU operators \\ (instances per second)\end{tabular}}} & \multicolumn{1}{c}{\textbf{\begin{tabular}[c]{@{}c@{}}Throughput of CPU operators with 1 core \\ (instances per second)\end{tabular}}} & \multicolumn{1}{c}{\textbf{Acceleration Ratio}} \\ \hline
Encoding & 15108620 & 77232 & 195 \\ \hline
Decoding & 22343405 & 587587 & 38 \\ \hline
Homomorphic Encryption & 83909 & 445 & 189 \\ \hline
Homomorphic Decryption & 287273 & 1465 & 196 \\ \hline
Homomorphic Multiplication & 991946 & 6410 & 155 \\ \hline
Homomorphic Addition & 1226803 & 27118 & 45 \\ \hline
Homomorphic Matrix Multiplication & 931942 & 4112 & 227 \\ \hline
Homomorphic Summation & 19575768 & 128893 & 152 \\ \hline
\end{tabular}%
}
\caption{Throughput of GPU HOs and CPU HOs with $10^5$ instances in each test. The bit-length of public key is 1024.}
\label{tab:Operator_Comparison}
\end{table*}

Through the processes above, redundant data transmissions are totally avoided. As GPU memory is limited, it is possible that GPU memory is not enough to cache all the calculation results. In our implementation, we leverage the LRU algorithm to select the cached results to be sent back to CPU and erased from GPU memory.

% strategy, memory constraint, context constraint (+ example)

\section{Preliminary Results}\label{sec:result}
\subsection{Implementation}
Our implementation mainly consists of two parts, including construction of GPU operators and \sys. The GPU operators are developed with CUDA Toolkit. \sys is implemented on open-sourced version of Federated AI Technology Enabler (FATE) 1.5.1~\cite{webank19}. We ran the framework on two CPU servers to simulate federated training process between two parties. The hardware configurations for both servers are the same, with CPU of \textit{Intel(R) Xeon(R) Silver 4114 CPU @ 2.20GHz}. We use a Nvidia Tesla V100 PCIe 32GB as the GPU accelerator for each server.

\subsection{Evaluation}
We conducted our experiments from two aspects. First, it is necessary to evaluate the performance of basic GPU operators because they are the fundamental components of the heterogeneous system. Second, in order to verify how well our system is compatible with FL frameworks, we train the logistic regression models with vanilla FATE and GPU-accelerated FATE separately.

The evaluation of operators is conducted by comparing the computing throughput between GPU and CPU. The CPU operators are implemented with \texttt{gmpy2} library, which is the same as that in FATE. Table~\ref{tab:Operator_Comparison} presents comparison results. The performance of GPU operators outperforms CPU significantly. Especially for operators like Homomorphic Multiplication which contains large amounts of modular exponentiation calculations, throughput of GPU operators is more than 100 times that of CPU operators.

To check the compatibility of \sys, we compare the performance of logistic regression model training with vanilla FATE and GPU-accelerated FATE. Model training is performed on credit card dataset\footnote{https://www.kaggle.com/arslanali4343/credit-card-cheating-detection-cccd}. We conducted vertical and horizontal partition on the dataset for heterogeneous LR and homogeneous LR respectively. 

\begin{figure}[!htbp]
    \centering
    \includegraphics[width=0.38\textwidth]{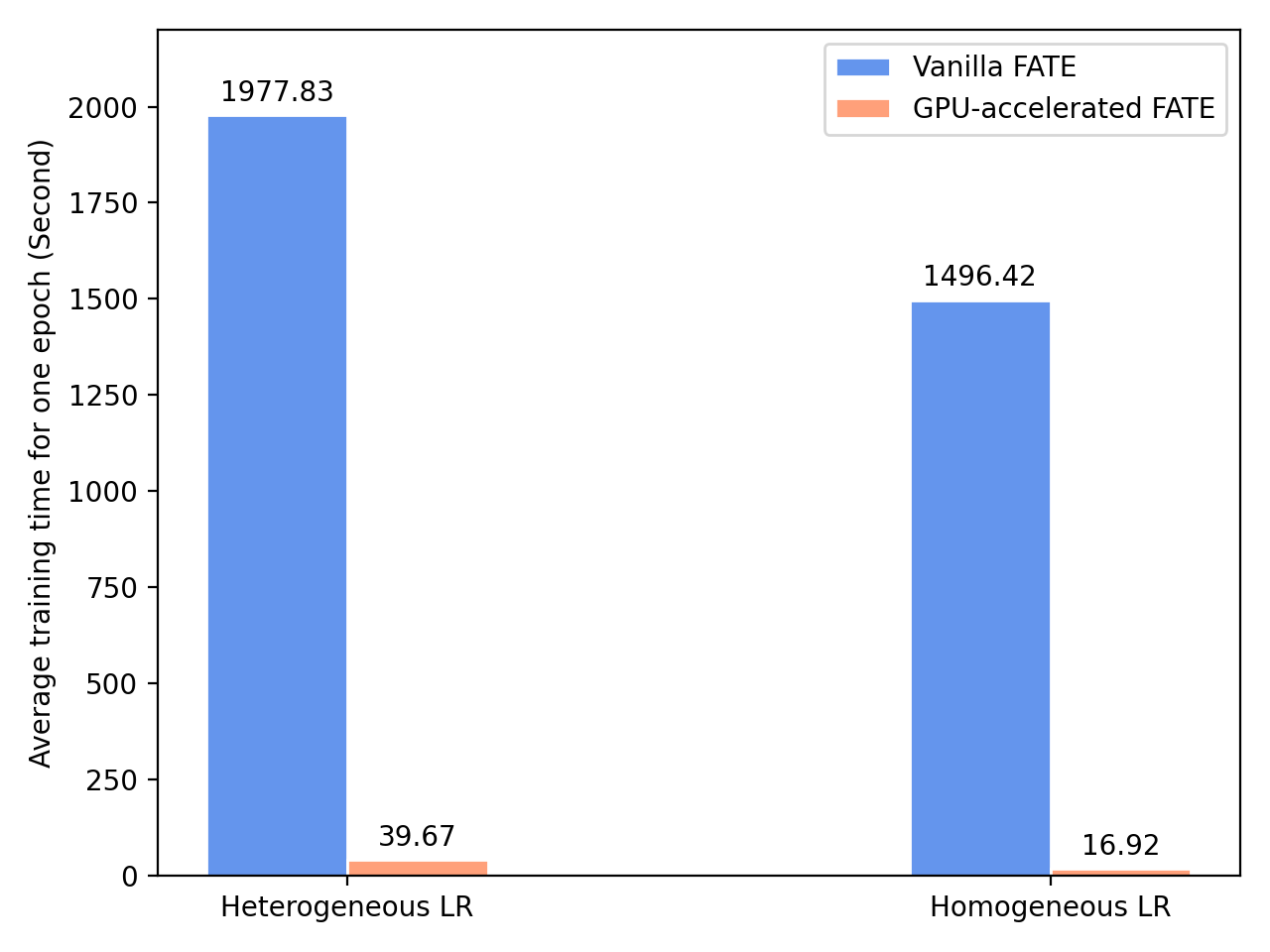}
    \caption{Performance of vanilla FATE and GPU-accelerated FATE}
    \label{eq:Performance}
\end{figure}

As shown in Figure~\ref{eq:Performance}, GPU-accelerated FATE has a remarkable performance increase for each training epoch compared to vanilla FATE. For heterogeneous LR, the modified FATE achieves an acceleration ratio of 49.9 compared with vanilla FATE in finishing one training epoch. For homogeneous LR, which is more computationally intensive, GPU-accelerated FATE achieves an acceleration ratio of 88.4.

\section{Conclusion and Future work}\label{sec:conclusion}
In this paper, we propose \sys, a GPU-based solution for accelerating FLR. To maximize the utilization of cross-device transmission bandwidth and GPU calculation resources, we optimize data storage with aggregation and serialization. Aimed at high computing throughput, several GPU-based homomorphic operators are implemented. Furthermore, an io-optimized workflow is introduced to minimize memory copy and cross-device data transfer by caching results in GPU memory. Based on our system, we accelerated an industrial FL framework and conducted logistic regression model training. The experimental results present the performance advantage of our solution.

As the need for FL in practical applications continues to grow, we have encountered some new challenges.

\emph{Needs for computational acceleration in various FL algorithms.} In addition to FLR, there are many other popular FL algorithms that suffer from the performance bottleneck of HE. According to the analysis, the performance degradation caused by data encryption is considerable in algorithms like SecureBoost and Federated Transfer Learning~\cite{jing2019quantifying}. Despite the progress in related research, it is still hard for new proposed algorithms to balance the trade-off between security and performance. Federated Matrix Factorization~\cite{chai2020secure} is a distributed framework which can be widely employed in scenarios like Federated Recommendation System~\cite{yang2020federated}. But more efficient encryption operations are urgently required to make the framework more practical in real-world applications. Therefore, it is essential to build a more comprehensive system. In the future, we will extend the functionality of \sys with more high-performance operators and generalized interfaces to support the acceleration for more FL algorithms.

\emph{Communication overhead in distributed system.} Due to large amounts of data and models, it is a common choice for a participant in FL to apply distributed system in data center network (DCN) to implement data storage and computation. Many FL frameworks use HTTP/2 based gRPC for data transmission, which may cause congestion in DCN. In some machine learning platforms like TensorFlow, end-to-end performance is possible to be improved by using RDMA to reduce network latency~\cite{yi2017towards}. We will explore the methods of combining GPU-accelerated FL and RDMA to reduce communication overhead while optimizing computation performance.

% \newpage
\bibliographystyle{named}
\bibliography{ijcai21}

\end{document}